\documentclass[11pt,british]{article} 

\usepackage{graphicx}
\usepackage{amsmath,amssymb} 
\usepackage{color}
\usepackage{multirow}
\usepackage{hyperref}
\usepackage{array}
\usepackage{soul}
\usepackage{babel}
\usepackage[margin=1.0in]{geometry}

\begin{document}

\title{Watermark Retrieval from 3D Printed Objects via Convolutional Neural Networks}

\author{
\parbox{1.0\textwidth}
{\centering
Xin Zhang, Qian Wang, Toby Breckon and Ioannis Ivrissimtzis\\
Durham University\\
Department of Computer Science\\
Durham, DH1 3LE, UK\\
\{xin.zhang3, qian.wang, toby.breckon, ioannis.ivrissimtzis\}@durham.ac.uk
}
}

\date{}

\maketitle

\begin{abstract} 
	We present a method for reading digital data embedded in planar 3D printed 
	surfaces. The data are organised in binary arrays and embedded as surface 
	textures in a way inspired by QR codes. At the core of the retrieval 
	method lies a Convolutional Neural Network, outputting a confidence map of 
	the location of the surface textures encoding value 1 bits. Subsequently, the bit 
	array is retrieved through a series of simple image processing and statistical operations 
	applied on the confidence map. Extensive experimentation with images captured from 
	various camera views, under various illumination conditions and from objects printed with 
	various material colours, shows that the proposed method generalizes well and 
	achieves the level of accuracy required in practical applications.
\end{abstract}

\maketitle

	
\section{Introduction}
\label{sec:introduction}


{\em Additive manufacturing}, most commonly known as {\em 3D printing}, 
is a manufacturing method that creates objects by adding one layer of material on top 
of the other. One of the advantages of additive manufacturing over more traditional 
methods is that it can handle geometries which either were impossible to manufacture, 
or the cost of their manufacturing was prohibitive \cite{thompson2016}. 


{\em Watermarking}, a term used in a broad sense in this paper, is the embedding of information 
on a physical object or a digital file. It may serve various purposes, such as object or 
file authentication, copy control, protection against unauthorized alteration, or 
dissemination of machine readable 
information. The latter is an increasingly popular application, especially in the 
form of QR codes, and is the target application of the proposed method for watermarking 3D printed 
objects. 

\begin{figure}
		\centering
		\includegraphics[width=0.24\linewidth]{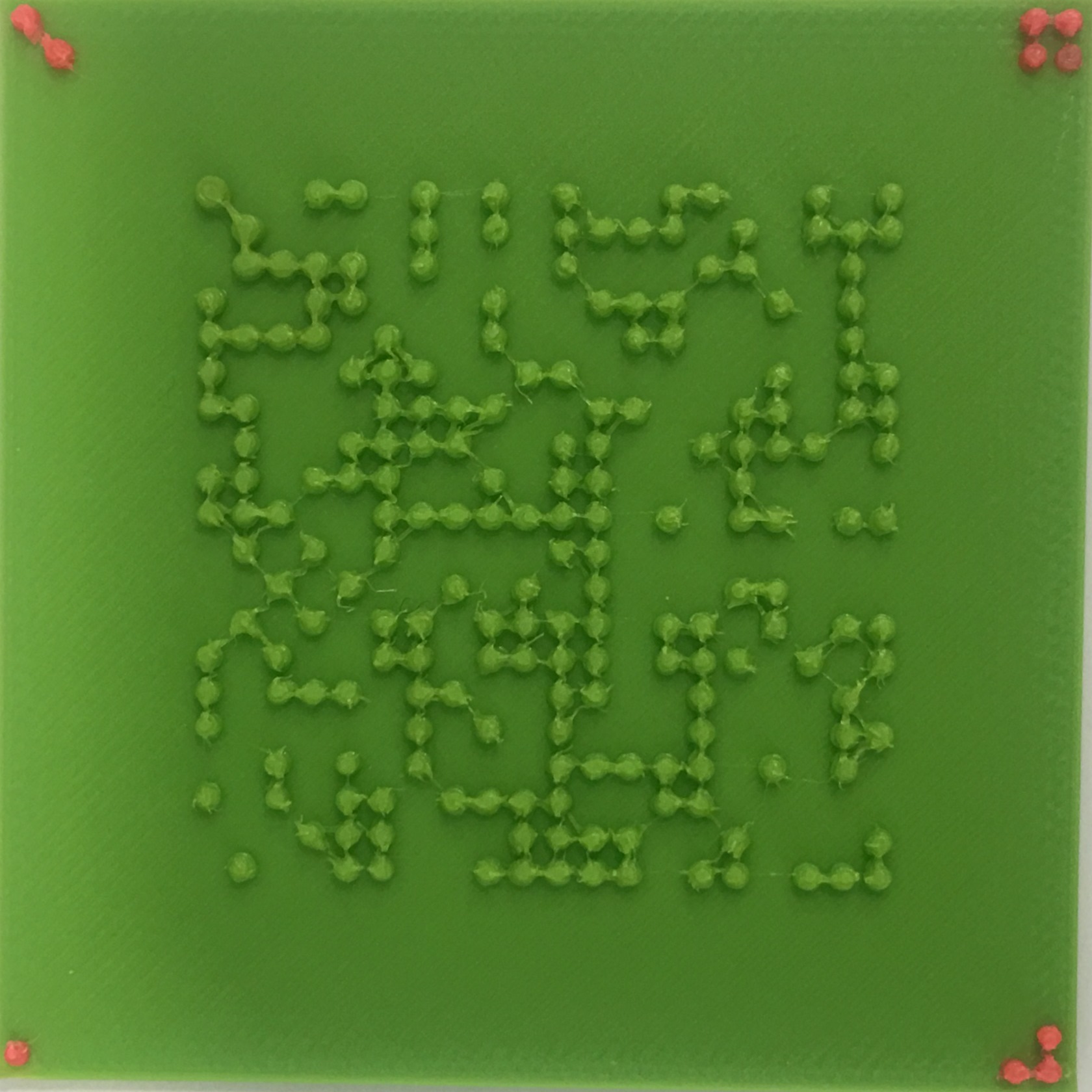} \hfill 
		\includegraphics[width=0.24\linewidth]{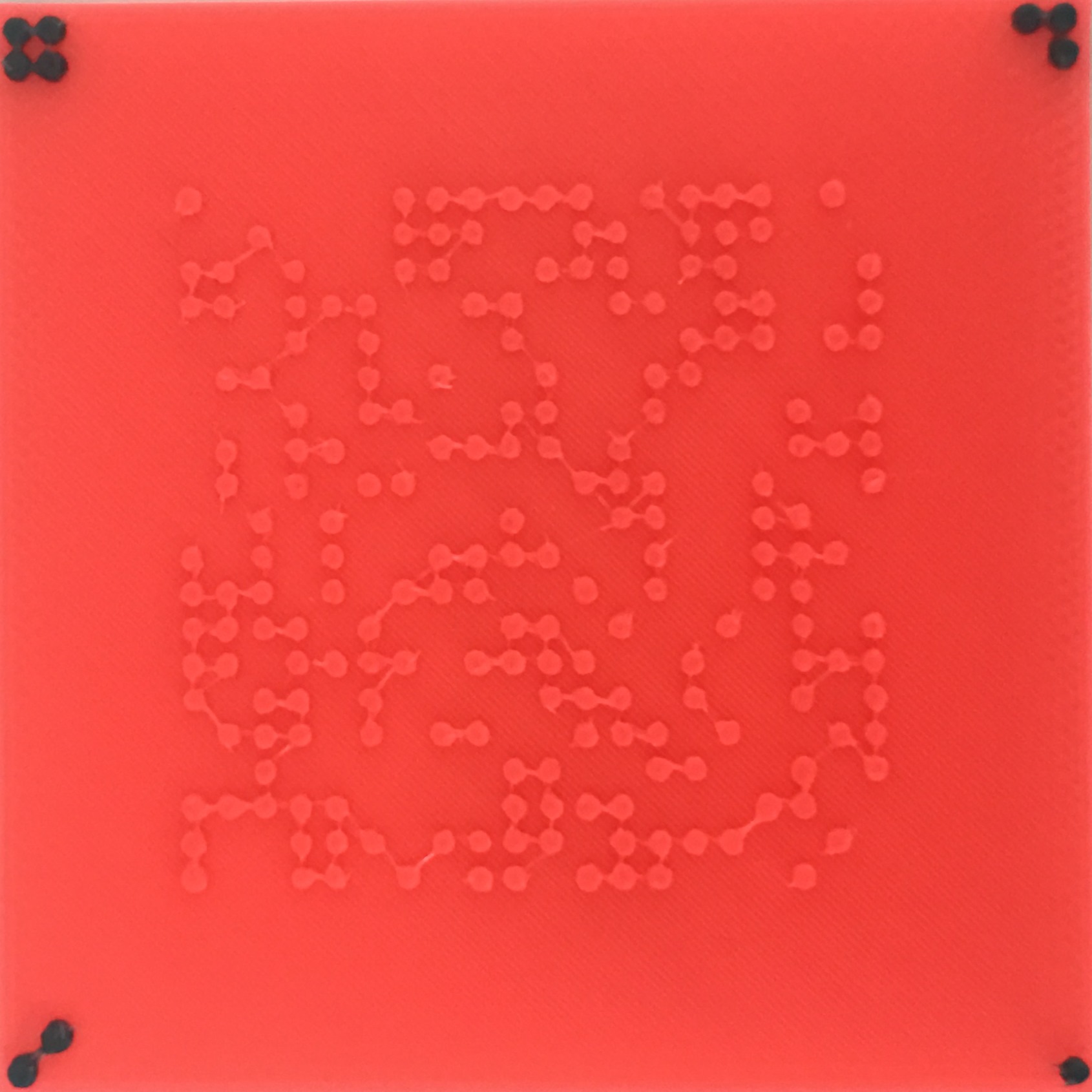}  \hfill 
	    \includegraphics[width=0.243\linewidth]{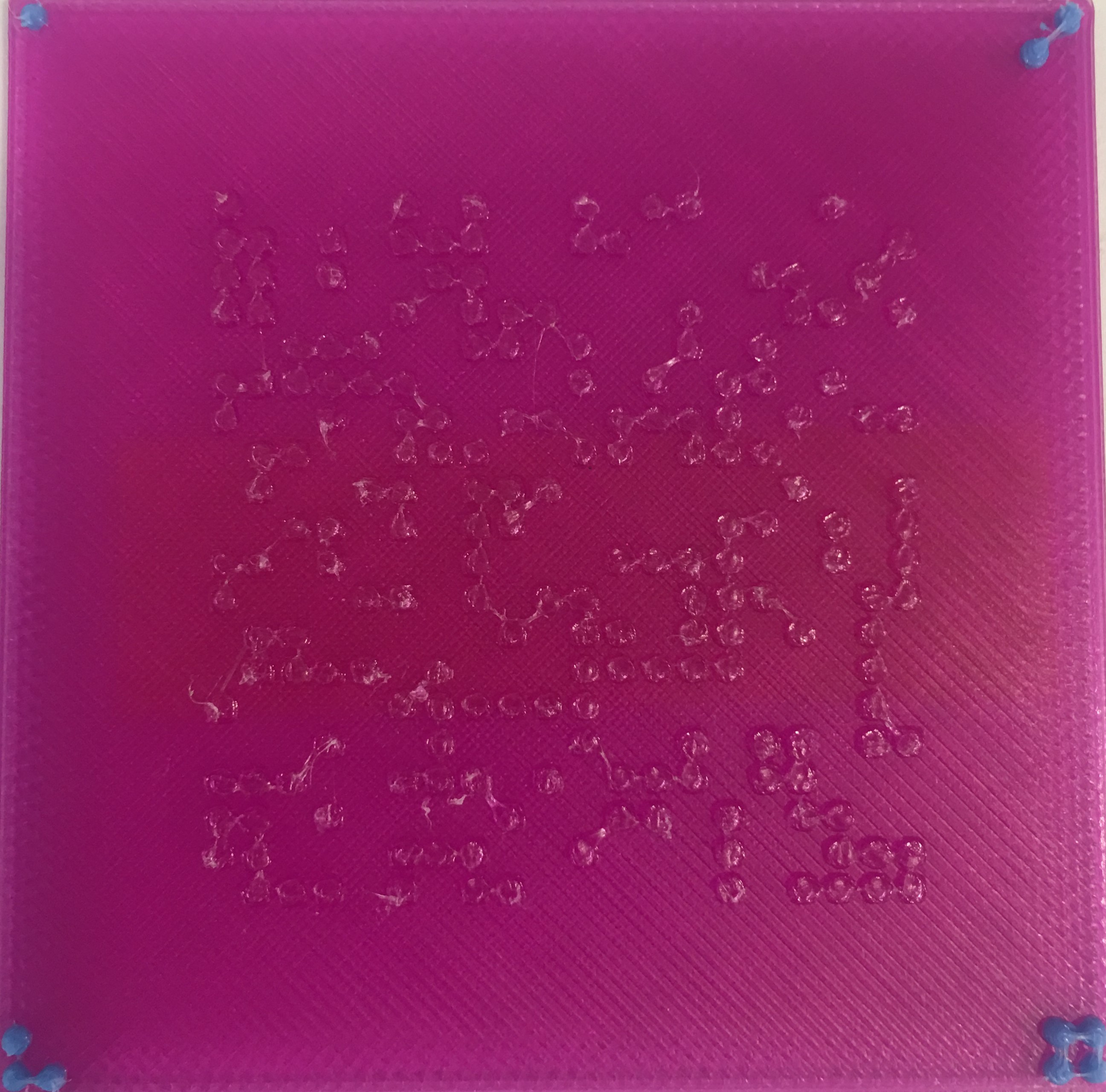} \hfill 
		\includegraphics[width=0.24\linewidth]{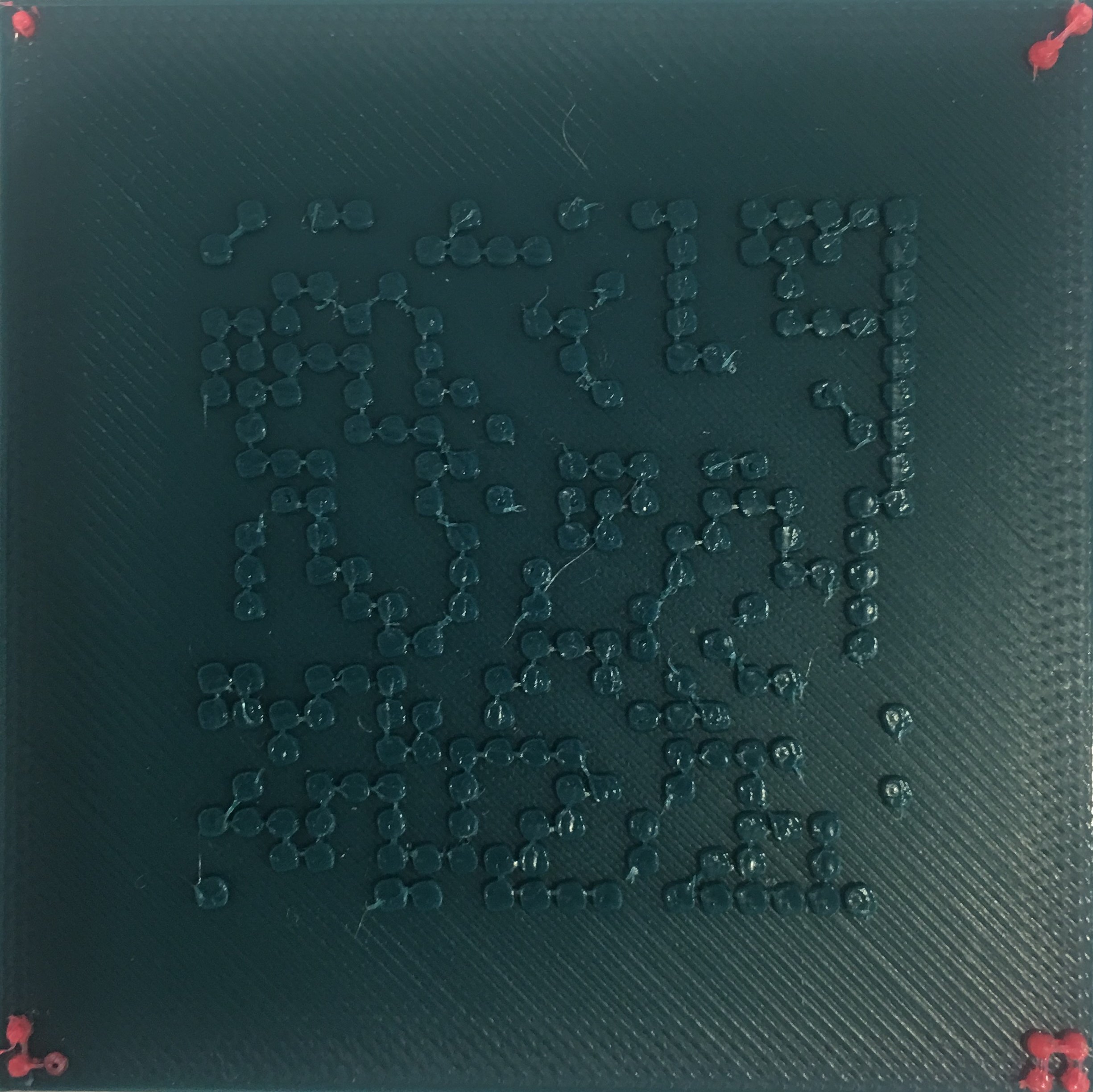} 
		\caption{Examples of 3D printed surface watermarks.}
		\label{fig:f0}
	\end{figure}

In this paper, aiming at exploiting the capability of 3D printing to 
create objects of more complex geometry at no additional cost, we propose a 3D surface 
watermark similar in design to QR codes. The information is embedded on the nodes of a 
regular grid, one bit per node, using semi-spherical or cubic bumps. The existence of a bump 
indicates a bit value of 1 at that node of the grid and its absence a bit value 0. 
See Figure \ref{fig:f0} for some examples. 


\subsection{Motivation} We notice that the common practice of placing a printed QR code 
on or next to a physical objects is vulnerable to various malicious attacks. For example, an 
attacker could replace a printed QR code with their own, or ovelay it, or in a phishing style 
attack put a printed QR code in a place where there shouldn't be one. The root cause of these  
vulnerabilities is that at the moment the manufacturing of an object and the embedding of 
information on it are two separate processes. 3D printing brings a unique opportunity to address 
the problem by integrating the information embedding process into the manufacturing process. That 
is, the embedded information can be part of the fabric of a 3D printed object, rather than been 
manufactured separately and glued on to it. 


\subsection{Challenges} 
\begin{figure}
	\centering
		\includegraphics[width=0.32\linewidth]{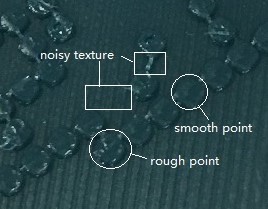} \hfill
		\includegraphics[width=0.32\linewidth]{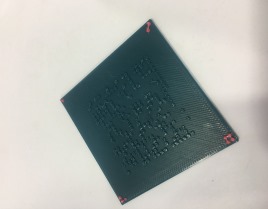} \hfill
    \includegraphics[width=0.32\linewidth]{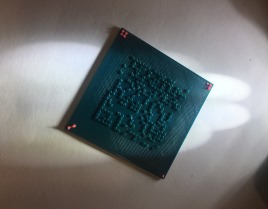}  
	\caption{{\bf Left:} Challenges related to the printing process. 
	{\bf Middle and Right:} Watermark retrieval challenges due to 
	illumination: natural light (middle) and uneven artificial light 
	(right).} 
	\label{fig:f1}
	\end{figure}

The challenges in developing a 3D printing watermark method can be 
summarized as follows: 
\begin{enumerate} 
	\item Since 3D printed objects are often created from a single material, the 
	background surface and the watermark carrying bumps would often have the same 
	colour and it would be difficult to distinguish between them. 
	\item The popularity of low-end printers with plastic filaments means that the carrier 
	surface would often contain significant noise as well as various infill patterns created 
	during slicing. For that reason, the sizes and shapes of the bumps may vary 
	considerably, even if they are identical in the digital file. See Figure \ref{fig:f1} (left). 
	\item Watermark retrieval is expected to be done from images captured by a 
	hand-held phone camera under potentially adverse conditions. Apart from 
	assuming an arbitrary camera viewpoint, the variability of the 
	lighting conditions, which may range from smooth natural light to extremely uneven artificial 
	illumination, should also be taken into account. See Figure \ref{fig:f1} (middle and right). 
\end{enumerate} 
Our experimental results show that convolutional neural networks can overcome this 
multitude of watermark retrieval challenges entailed into a realistic scenario.

\subsection{Contributions} 
\begin{figure}
	\centering
	\includegraphics[width=\linewidth]{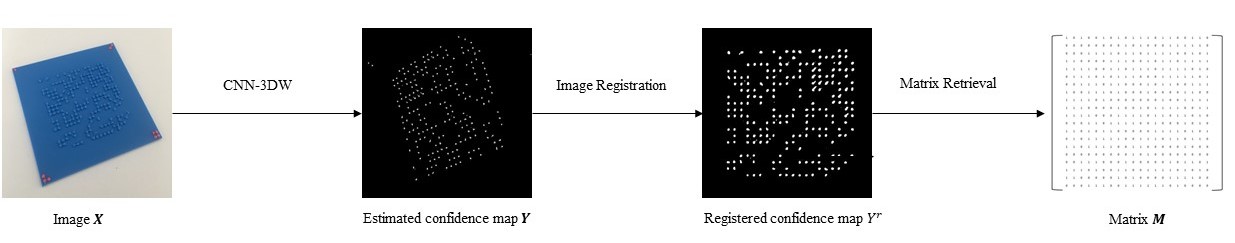}
	\caption{The pipeline of the proposed 3D watermarking retrieval method.}
	\label{fig:zero}
\end{figure}

The main contributions of the paper are the development of a 
watermark retrieval algorithm the design and creation of an image database of 3D printed 
watermarked objects. The retrieval algorithm starts with a convolutional neural network 
generating an approximate density map representing the locations of watermark bumps 
in the input RGB image. The subsequent steps are a series of simple image processing and 
statistical operations. Specifically, the registration of the density map, a sequence of simple 
image processing operations, and finally, the retrieval of the watermark through the K-means 
clustering method. See Figure \ref{fig:zero} for a high level description of the algorithm. 

	\section{Related Work}
\label{sec:related}

Comprehensive reviews of additive manufacturing with academic or industrial focus 
can be found in \cite{wong2012review,terry2012additive}. Currently, there is 
a multitude of competing additive manufacturing technologies, from laminated object 
manufacturing to photosensitive polymer curing molding to laser powder sintering 
molding. All our test objects were printed with fused deposition modeling, which 
is a low-cost and thus widely available 3D printing technology. 

\subsection{Digital 3D Watermarking}

Digital 3D watermarking is an active, well-developed field 
\cite{ohbuchi02,bors2006,luo11,yang17}, mostly focusing on boundary surface 
representations of 3D models, triangle meshes in particular. 
The proposed watermarking of 3D printed surfaces has some obvious relationships 
with digital 3D watermarking. Indeed, the construction of a watermarked 3D 
printed object starts with the creation of a watermarked 3D file. However, 
the two processes diverge significantly during watermark retrieval. In digital 
3D watermarking, we have to extract information from a digital file by 
inverting the embedding algorithm. In 3D printing watermarking in contrast, 
having to extract information from a physical object we are faced with a 
classic computer vision problem.  

The challenge in the development of digital 3D watermarking algorithms is to 
make them robust against a variety of malicious or unintentional attacks, 
including 3D printing and rescanning. Watermark survivability under this 
attack has been tested in \cite{macq15}, while surface watermarking algorithms 
resilient to the attack have been proposed in \cite{hou15}. Notice that the aim 
of the watermarking algorithm in \cite{hou15} is to protect the digital 3D file 
from watermark removal attacks while here our aim to embed machine readable 
information on physical objects. Guided by different objectives, they restrict 
themselves to just 24 bits of payload and most importantly, retrieval
involves 3D laser scanning and surface reconstruction. In contrast, we do not 
make use of any specialized equipment such as laser scanners as we never
reconstruct a digital 3D model from the 3D printed object.

\subsection{Feature extraction} 

Convolutional neural networks have demonstrated impressive performance on a variety of 
computer vision tasks \cite{lecun2015deep}\cite{lecun2010convolutional}\cite{robert2014machine}. 
They are considered particularly well suited for feature 
extraction and classification tasks, AlexNet \cite{krizhevsky2012imagenet}, 
VGGNet \cite{simonyan2014very} and GoolgeNet \cite{szegedy2015going} being some 
famous examples. Recently, deep CNN models have been employed for image based crowd counting tasks \cite{zhang2015cross}\cite{zhang2016single}\cite{mundhenk2016large}. It is usually formulated as a regression problem with the raw images as input and outputting a feature map (i.e. density map) characterizing the density of crowd in the image. Here, following,
we take the simple but effective approach of training a CNN to 
extract a confidence map. 

In an earlier approach to the problem of 3D printed watermark retrieval we developed 
a technique for recognizing the watermark bumps based on local binary patterns (LBP) 
\cite{ojala2000gray}. However, as it is often the case with hand-crafted feature 
extraction methods, the results did not generalize very well under adverse conditions, 
in particular when the background patterns were too prominent, or under extreme uneven 
illumination, or under unfavorable camera viewpoints. 
	
	\section{Method} 
\label{sec:method} 

Given an image/photo of a printed watermark as shown in Figure \ref{fig:f1}, we aim to extract the information embedded in the watermark, which in our case is a binary matrix. To detect the surface bumps corresponding to bit value 1 we propose an image processing pipeline comprising of three stages (c.f. Figure \ref{fig:zero}). In the first stage, a fully covolutional network (FCN) takes an image $\mathbf{X}$ as input and outputs an estimated confidence map $\mathbf{Y} = f(\mathbf{X})$. To overcome the issues of target scaling, rotation and affine distortion in the input images, in the second stage, image registration procedures are applied to the confidence map $\mathbf{Y}$ to get a registered confidence map $\mathbf{Y}^r$. Finally, the information matrix $\mathbf{M}$ is extracted from the registered confidence map $\mathbf{Y}^r$. In the following subsections, we explain these three stages in detail.
	
\subsection{CNN-3DW}
\label{sec:fcn}
\subsubsection{Network Architecture}
\begin{figure}
	\centering
	\includegraphics[width=\linewidth]{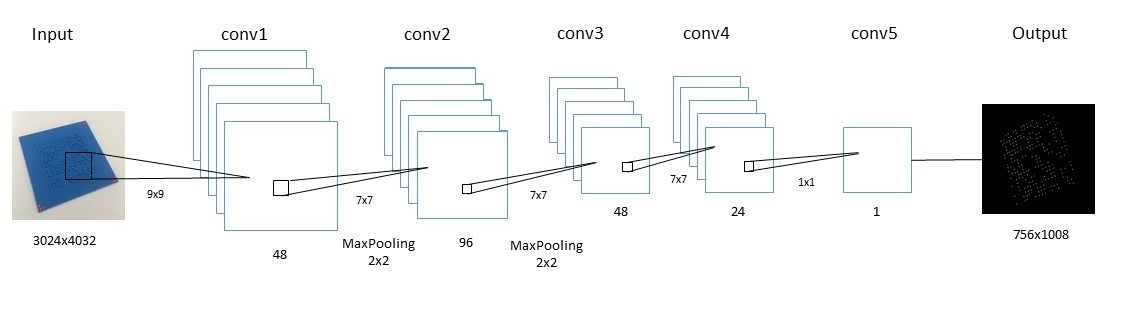}
	\caption{The architecture of the proposed CNN-3DW.}
	\label{fig:fourth}
\end{figure}

The proposed CNN based model for 3D watermark retrieval is a fully convolutional network named \textbf{CNN-3DW}. The architecture is shown in Figure \ref{fig:fourth}. It consists of five convolutional layers in which there are 48, 96, 48, 24, 1 kernels with sizes of $9\times9, 7\times7, 7\times7,7\times7,7\times7$ and $1\times1$ respectively. The activation function is the Rectified Linear unit (ReLu), which generally performs well in CNNs, see for example \cite{zeiler2013rectified}. Two max pooling layers with step size two are added after the first and second convolutional layers, respectively. For all the convolutional layers, we set the step size as one and do zero padding so that the image size is not altered by the convolution operations. As a result, the final output has one fourth the size of the input image due to two maxpooling layers. Since the whole network is fully convolutional, it is able to handle input images of different sizes.

\subsubsection{Ground Truth Confidence Map}
\begin{figure}
		\centering
		\includegraphics[width=0.3\linewidth]{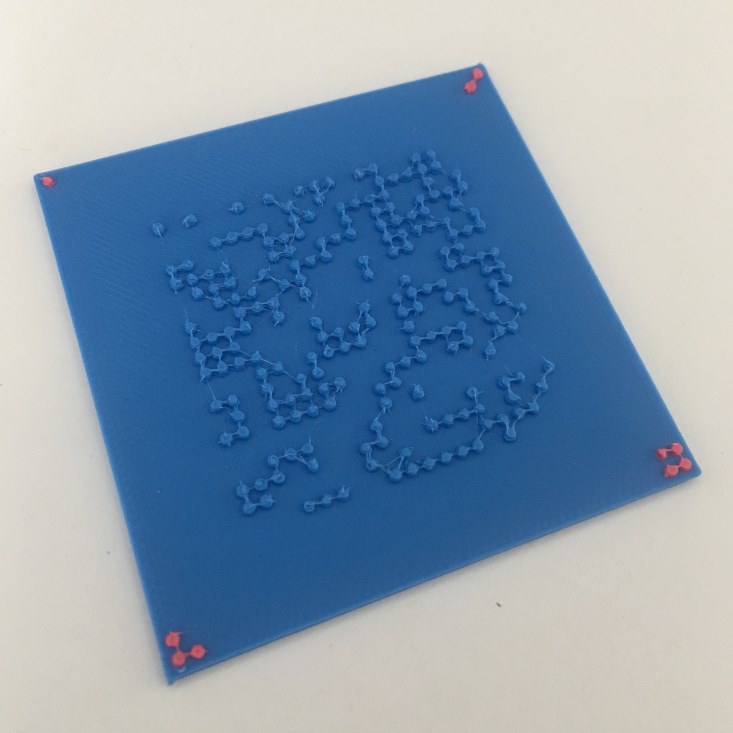} \hfill 
		\includegraphics[width=0.3\linewidth]{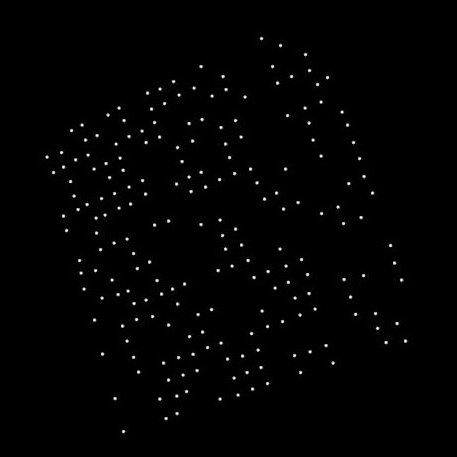} \hfill 
		\includegraphics[width=0.3\linewidth]{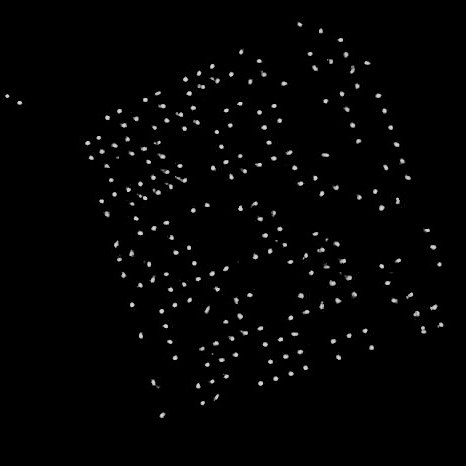}
		\caption{\textbf{Left:} image of a watermarked object. \textbf{Middle:} ground truth map.\textbf{Right:} corresponding confidence map.}
		\label{fig:fifth}
	\end{figure}

Given an input image $\mathbf{X}$ of a printed watermark, the goal of our CNN-3DW is to estimate the confidence map of ``bump" locations representing the value of ones in the information matrix. Figure~\ref{fig:fifth} shows an image of a watermarked object and its corresponding ground truth confidence map and the estimated confidence map by the proposed CNN model.
 We need to annotate the training data for such purpose. Specifically, we generate the ground truth confidence map $\mathbf{Y}$ based on the annotation image $\mathbf{A}$ by a Gaussian smoothing process.
\begin{equation}
\label{equ:gaussian}
\mathbf{Y} = \mathbf{A}*\mathbf{G}_{\sigma},
\end{equation}
where $*$ is a convolution operator applied on the binary annotation image $\mathbf{A}$ and the Gaussian kernel $\mathbf{G}_{\sigma}$. The annotation image can be represented by
\begin{equation}
\mathbf{A}_{ij} = 
\begin{cases}
1 \quad \text{if $(i,j)$ is the centroid of a marked region,}
\\
0 \quad \text{otherwise}.
\end{cases}
\label{equ:one}
\end{equation} 
The annotation images are the results of human annotation on the training data while the standard deviation of Gaussian kernel was determined empirically to be $\sigma=5$.
\subsubsection{Model Training}
To train the network, we use the mean squared error (MSE) loss function:
\begin{equation}
L(\Theta)=\frac{1}{N\times H\times W} \sum_{n,h,w}(f(\mathbf{X}; \Theta)-\mathbf{Y})^{2},
\label{equ:three}
\end{equation}
where $\Theta$ is the set of network parameters; $N$ is the number of 
training images;  $\mathbf{X}$ is an input image, $\mathbf{Y}_{i}$ the ground truth confidence map of $\mathbf{X}$ and $f(\mathbf{X},\Theta))$ is the predicted confidence map of the input image $X$; $W$ and $H$ are the width and height of the confidence map image $\mathbf{Y}$ respectively. 

Data augmentation techniques have proved to be beneficial to the training of deep CNN models, alleviating the overfitting issue especially in situations where the training data is not large enough. In our experiments we employ three data augmentation techniques to compensate for insufficient training data. Firstly, the images are randomly rotated by a random angle degrees. Secondly, the training data are augmented by randomly changing their average brightness. Finally, random cropping is applied to generate 10 patches of uniform size (i.e. $512\times 512$) from each image. 

We trained the CNN-3DW with Keras \cite{chollet2015keras} built on TensorFlow. The Adam optimizer \cite{kingma2014adam}, a stochastic gradient descent optimization method for training deep learning models, is employed with the default parameter values.
We set the initial learning rate as 1\(e -5\) and decrease it to 1\(e -6\) after 50 epochs. The training is stopped after 100 epochs where we observed convergence of the model. When using an NVIDIA GeForce GTX TITAN X, the average training time was about 7 hours on our own dataset which will be introduced in the following section.\par

\subsection{Image registration, processing and matrix retrieval} 
Using the trained CNN-3DW model we are able to estimate a confidence map $\mathbf{Y}$ of the watermark bumps in the image $\mathbf{X}$. To extract the embedded information matrix from the estimated confidence map $\mathbf{Y}$ is still non-trivial. The first issue is that we need to locate the watermark regions in the confidence map. In our experiments, we observed that noises exist in the background regions of some images, posing great challenges to the automatic watermark region localization. Significant bias in the watermark region localization would lead to catastrophic error in the following stages. To fight this issue off, we print four \textit{landmarks} at the corners of the watermark with easily distinguishable colours (see Figure \ref{fig:f0}). During retrieval, the watermark regions are easily located by finding the ``differently coloured" \textit{landmarks} at the four corners. 

The second issue is that the watermark in the image could be of variant scales, rotations and affine distortions. To tackle this issue, we use image registration \cite{brown1992survey} to transform the watermark region in the confidence map $\mathbf{Y}$ to a square region (see Figure \ref{fig:zero}). In image registration, the \textit{landmarks} are used as control points which are supposed to be mapped to the four corners of the new square region. In our experiments, we used Matlab's built-in tools for the image registration. After image registration, we obtain the transformed confidence map and the watermark region is cropped with the background region discarded. We denote the reserved watermark region of confidence map as $\mathbf{Y}^r$.

\begin{figure}[h]
			\centering
		\includegraphics[width=0.33\linewidth]{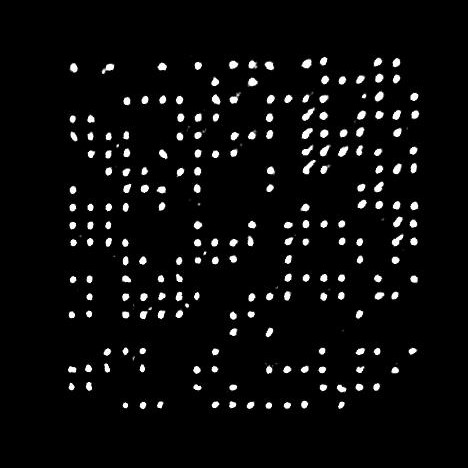} \hspace{1cm}
		\includegraphics[width=0.33\linewidth]{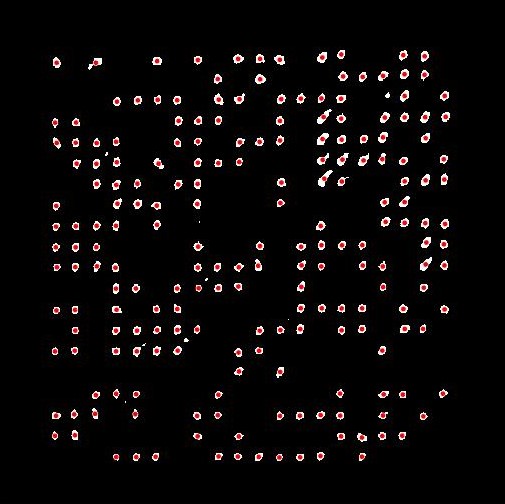}
		\caption{The regularised CNN-3DW output (left) and the detected centroids of bit value 1 regions (right).}
		\label{fig:seventh}
	\end{figure}

Finally, we extract the embedded information matrix $\mathbf{M}$ from $\mathbf{Y}^r$. The registered confidence map $\mathbf{Y}^r$ is firstly binarized by a threshold $t$,
\begin{equation}
\label{equ:binarize}
\hat{\mathbf{Y}}^r_{ij} =
\begin{cases}
1 \quad \text{if $\mathbf{Y}^r_{ij} > t$ ,}
\\
0 \quad \text{otherwise}.
\end{cases}
\end{equation}
where $t =\beta\times Thre$ and $Thre$ is the OTSU threshold \cite{otsu1979threshold} of the registered confidence map $\mathbf{Y}^r$, $\beta$ is a user defined parameter. In our experiments we empirically set $\beta=0.35$ for optimised performance. The binary confidence map $\hat{\mathbf{Y}}^r$ is visualized in Figure \ref{fig:seventh}. We call Matlab's {\em regionprops} function to detect the connected regions within the binary image and obtain estimates of their centroids and two semi-axes. If the the sum of the two semi-axes is above a threshold, a bit value 1 will be returned from the centroid of that region as shown in Figure~\ref{fig:seventh}. Now we end up with a set of coordinates representing the detected watermark bumps in $\mathbf{\hat{Y}}^r$. As a final step, we extract the information matrix $M$ from the set of coordinates. Ideally, there exist only $m$ (the number of rows and columns of $\mathbf{M}$) unique values for $x$ and $y$ coordinates, however, due to the imperfection of the results of the previous steps, the coordinates have biases. To resolve this issue, we utilize a $K$-means clustering on the $x$ and $y$ coordinates of all the detected points, respectively. We set the number of clusters as $m$ and rank the $m$ cluster centers. The point falling into the $i$-th cluster will be assigned to a new coordinate in the $m\times m$ matrix $\mathbf{M}$.



	
	\section{Experiments} 
\label{sec:experiments} 
In this section, we conduct a series of experiments to evaluate the effectiveness of the proposed approach on 3D watermark retrieval. Since there are no datasets with images of 3D printed objects watermarked with a similar method, we designed our own dataset named \textbf{3DW-FS} \footnote{The whole dataset including raw images and the annotations will be made publicly available.}. We conduct five experiments and report the results in the following subsections.

\subsection{Dataset: 3DW-FS}
\begin{figure}
	\centering
	\includegraphics[width=0.4\linewidth]{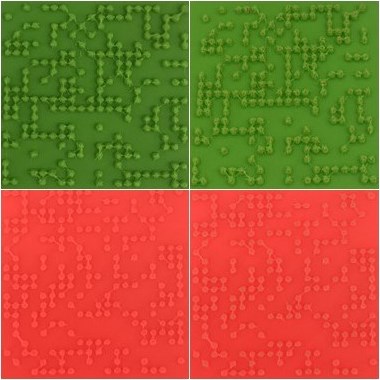}
	\caption{Two example of raw images (left) and annotated images (right) for ground truth bump location extraction.}
	\label{fig:datasetTrain}
\end{figure}

To collect a dataset as the test bed for our validation experiments, we first print watermarks on multiple objects, then collect images of the watermarked objects for our experiments. The watermarks printed on objects convey information encoded in $20\times 20$ binary matrices which are randomly generated and different for each object. The images are captured using a mobile phone camera which takes images of $3024 \times 4032$ resolution. The dataset consists of 290 images from 20 objects printed with 9 different materials of different colour. We collect 15 images for each of the objects 1-16 and 5 images for each of objects 17-20. Figure \ref{fig:datasetTrain} shows one exemplar image from each printed object. 
\begin{figure}
	\centering
	\includegraphics[width=0.5\linewidth]{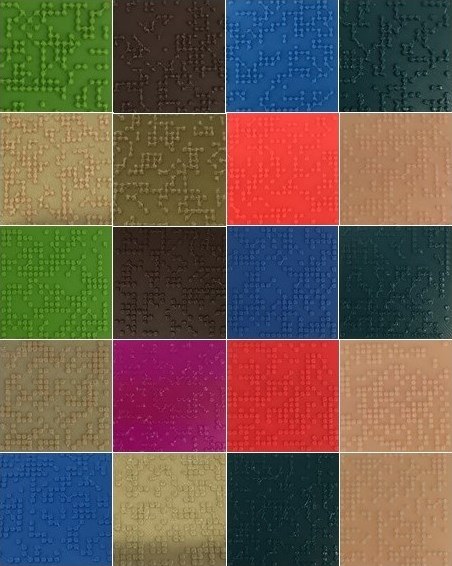} 
	\caption{Exemplar images of objects 1-20 from left to right and top to down.}
	\label{fig:datasetAll}
\end{figure}
To simulate potential scenarios in practical applications, we captured images under variant conditions. Firstly, we collected 10 and 5 images under natural and extreme artificial illumination respectively for objects 1-16; while for objects 17-20, we collected 3 and 2 images under these two illumination conditional respectively. Secondly, we keep variant distances to the objects when collecting images, so that the watermark regions in the images have different sizes. Thirdly, we ensure that the watermarks in the collected images have variant rotations in both horizontal and vertical directions to expect that the watermark can be recognized from arbitrary perspectives. As a result, our dataset is challenging and provides a simulation of real scenarios.

For the purpose of CNN model training, we manually annotate all the images of objects 1-16, i.e., 240 images in total. For the annotation, we simply put a dot at the center of a watermark bump in the image as shown in Figure \ref{fig:datasetTrain} from which the ground truth bump locations can be easily extracted. The ground truth targets of the CNN model are then computed based on the annotated bump locations by Eq.(\ref{equ:gaussian}).

\subsection{Performance measures} 
Generally, the retrieval of watermark bits can be 
seen as a binary classification problem with two symmetric classes. However, since in 
our case value 0 bits correspond to no alteration of the carrier surface and are 
indistinguishable from the background, we treat the retrieval process as a binary 
detection test. Thus, as performance measures of the overall watermark retrieval method 
we report sensitivity TPR=TP/(TP+FN), specificity SPC=TN/(TN+FP), precision PPV=TP/(TP+FP) 
and negative predictive value NPV=TN/(TN+FN), where TP and FN are the numbers of value 1 
bits retrieved correctly and correspondingly incorrectly, while TN and FP are the 
numbers of value 0 bits retrieved correctly and correspondingly incorrectly.  

\subsection{Experimental Results} 

We report results from four tests. 
\begin{enumerate}
	\item A 5-fold cross-validation test on the 16 printed objects that were used for the development of the CNN-3DW. 
	\item A validation test on images from four objects printed after the development of the CNN-3DW was complete. 
	\item A cross-material validation test where training and test sets do not contain images of objects printed with the same material colour. 
	\item A test demonstrating the feasibility of an active learning approach, in which we identify a material colour on which the  
	algorithm performs poorly, print more objects using that material and use their images to retrained the network. 
\end {enumerate}

\subsubsection{Experiment 1}
In this experiment we assess the general accuracy of the method and how it is 
affected by material colour and illumination conditions. We use objects 1-16 which are printed using 
9 material colours in total. From each object we capture 15 images of size 
$3024\times 4032$ with an iPhone 8 camera under different, arbitrary perspectives, 
10 of them under natural light and 5 under extreme artificial illumination. 
From each object we select two natural and one artificial illumination image to create a 
test set containing 48 images in total, while the other 192 images used for training. We repeat 
this process 5 times, until all images have been used once for testing. 

\begin{figure*}
		\centering
		\includegraphics[width=0.33\textwidth]{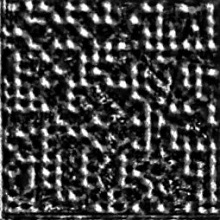} \hspace{1cm}
		\includegraphics[width=0.33\textwidth]{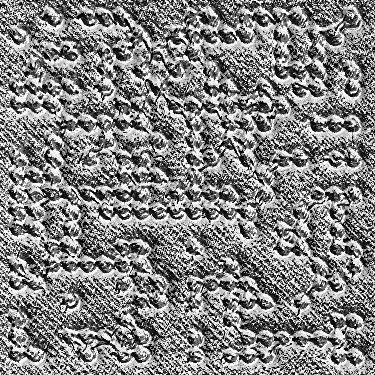} 
		\caption{LBPs of the Wooden (left) and Green (right) material printouts.}
		\label{fig:rebuttal}
	\end{figure*}

The results are summarized in Table~\ref{tbl:t1}. We can see that the proposed method performs well 
on most of the printed objects, dark green and transparent purple being the most notable exceptions. 
In contrast, a traditional image processing and analysis pipeline based on Local Binary Patterns and 
ellipse detection, which we developed for comparison purposes, did not generalize well. 
Figure \ref{fig:rebuttal} shows the LBPs from Wooden and Green material printouts. On the low 
background noise Wooden printout, they outperform CNNs with retrieval rates above 90\%, however, on
the high background noise Green material the traditional pipeline breaks down. 

Notice that in this experiment we used only one transparent purple object, thus the poor material 
properties and not enough training data could explain the poor retrieval performance. In experiment 
4 we employ a form of active learning \cite{lecun2015deep} to study this further. 

Table~\ref{tbl:t2} shows average performances over all 16 printed objects for each illumination
condition. As expected, extreme artificial light has a negative effect on performance. 

\begin{table}[h]
	\centering
			\caption{Experiment 1: 5-fold cross validation.} 
	\begin{tabular}{c|l||m{1.0cm}|m{1.0cm}|m{1.0cm}|m{1.0cm}}
		\hline
		& \textbf{Colour} & TPR & SPC & PPV & NPV \\
		\hline
		1 & Green (a) & 0.965 & 0.966 & 0.967 & 0.965 \\ 
		2 & Green (b) & 0.994 & 1 & 1 & 0.993 \\ 
		3 & Dark Brown (a) & 0.798 & 0.863 & 0.854 & 0.812 \\ 
		4 & Dark Brown (b) & 0.905 & 0.940 & 0.937 & 0.912 \\
		5 & Blue (a) & 0.964 & 0.998 & 0.998 & 0.966 \\
		6 & Blue (b) & 1 & 0.998 & 0.998 & 1 \\
		7 & Dark Green (a) & 0.662 & 0.698 & 0.699 & 0.661 \\
		8 & Dark Green (b) & 0.882 & 0.942 & 0.939 & 0.887 \\	
		9 & Wooden (a) & 0.830 & 0.840 & 0.851 & 0.818 \\
		10 & Wooden (b) & 0.862 & 0.881 & 0.871 & 0.872 \\
		11 & Luminous Red (a) & 0.971 & 0.981 & 0.982 & 0.970 \\
		12 & Luminous Red (b) & 1 & 1 & 1 & 1 \\ 
		13 & Skin (a) & 1 & 1 & 1 & 1 \\
		14 & Skin (b) & 0.882 & 0.921 & 0.914 & 0.893 \\
		15 & Bronze & 0.889 & 0.901 & 0.900 & 0.890 \\
		16 & Transparent Purple & 0.542 & 0.862 & 0.756 & 0.685 \\
		\hline 
		& &  \textbf{0.895} & \textbf{0.897} & \textbf{0.895} & \textbf{0.885} \\
		\hline
	\end{tabular}
	\label{tbl:t1}

	\vskip 0.5cm
		
	\centering
	\caption{Performance comparison under different illumination conditions.}
	\begin{tabular}{r||p{1cm}|p{1cm}|p{1cm}|p{1cm}}			
	\hline
	& TPR & SPC & PPV & NPV \\
	\hline
	Natural light & 0.89 & 0.93 & 0.92 & 0.89 \\
	Extreme artificial light & 0.82 & 0.86 & 0.85 & 0.84 \\
	\hline
 \end{tabular}
\label{tbl:t2}	
\end{table}

\subsubsection{Experiment 2} 

For this experiment we use images of objects 1-16 for training and test with images of objects 17-20. Notice that while in the previous experiment the watermark matrices of the test images also appear in the training set, with this test we can exclude the remote chance that CNN-3DW is memorizing parts of watermark matrices. 

The results are shown in Table \ref{tbl:t3}. Apart from averages we also report the results 
of an ensemble algorithm determining the value of each bit by a majority vote over the five 
images. Notice that the ensemble method is very much within the spirit of our target 
application scenario since modern mobile phones can easily capture and process short video 
sequences. The results show that CNN-3DW retrieval rates are independent if the watermark 
matrix and that majority vote can improve retrieval accuracy to levels that would be deemed 
satisfactory in a real life application. 
	\begin{table}[h]
		\centering
		\caption{Test results on four post-development printed objects.}
		\label{tbl:t3}
		\begin{tabular}{p{1.2cm}|l||p{1cm}|p{1cm}|p{1cm}|p{1cm}}							
	\hline	
	Colour & Metrics & TPR & SPC & PPV & NPV \\
	\hline	
	\multirow{2}{1em}{Blue}&Average& 0.99&1.00&1.00&0.99\\
	&Majority Vote&1.00&1.00&1.00&1.00\\
	\hline
	\multirow{2}{1em}{Wooden}&Average&0.87 &0.86&0.85&0.88\\
	&Majority Vote&1.00&1.00&1.00&1.00\\
	\hline
	\multirow{2}{1em}{Dark Green}&Average&0.81&0.85&0.84&0.82\\
	&Majority Vote&0.98&1.00& 1.00& 0.98\\
	\hline
	\multirow{2}{1em}{Skin}&Average&0.93&0.99&0.99&0.94\\
	&Majority Vote&1.00&1.00& 1.00& 1.00\\
	\hline
\end{tabular}
\end{table}

\subsubsection{Experiment 3}

Here we use the same test set as in Experiment 2, but each of the 
four newly printed test objects is tested on a neural network 
trained only with images of different material colours. The aim 
of this test is to assess the generalisation ability of CNN-3DW on 
images from objects with unseen material colours. 

The results are shown in Table~\ref{tbl:t4}. From the4 comparison between the results 
of Tables~\ref{tbl:t3} and \ref{tbl:t4} we notice that CNN-3DW can handle some 
colours such as Blue and Dark Green quite well as unseen colours, while in some other 
cases (such as Wooden and Skin) the performance drops significantly. 
	\begin{table}
	\centering
	\caption{Test results on objects printed with unseen material colours.}
	\label{tbl:t4}
		\begin{tabular}{p{1.2cm}|l||p{1cm}|p{1cm}|p{1cm}|p{1cm}}						
		\hline	
		Colour & Metrics & TPR & SPC & PPV & NPV \\
		\hline	
		\multirow{2}{1em}{Blue} & Average & 0.95 & 0.97 & 0.97 & 0.95 \\
		& Majority Vote & 1.00 & 1.00 & 1.00 & 1.00 \\
		\hline
		\multirow{2}{1em}{Wooden} & Average & 0.79 & 0.76 & 0.76 & 0.79 \\
		& Majority Vote & 0.87 & 0.83 & 0.82 & 0.87 \\
		\hline
		\multirow{2}{1em}{Dark Green} & Average & 0.74 & 0.82 & 0.80 & 0.76 \\
		& Majority Vote & 0.91 & 0.94 & 0.94 & 0.91 \\
		\hline
		\multirow{2}{1em}{Skin} & Average & 0.65 & 0.8 & 0.81 & 0.74 \\
		& Majority Vote & 0.72 & 0.98 & 0.98 & 0.76 \\
		\hline
	\end{tabular}
	\end{table}

\subsubsection{Experiment 4}

3D printed watermarking retrieval is an application well suited for the adoption 
of an active learning approach where we expand the training set with examples 
on which the current network underperforms and then we retrain. Indeed, on 
one hand, performance assessment of the ability of the algorithm to retrieve 
the correct bit values can be done completely automatically, on the other hand, 
the expansion  of the training set requires the hand-annotation of images of 
watermarked objects, which is a laborious, tedious process. Therefore, being 
able to choose suitable images for inclusion to the training set is a matter of 
importance. 

From Table~\ref{tbl:t1} we notice that transparent purple is a particularly 
challenging material colour. We printed two more watermarked objects on this 
colour, capture 15 images from each object and added them to the training set.  
From Table~\ref{tbl:t5} shows the performance of the algorithm, before and 
after retraining with the expanded set, on three the three transparent purple 
test images where it initially performed the worst. We notice a considerable 
performance improvement. 

	\begin{table}
	\centering
	\caption{An active learning approach on transparent purple objects.}
	\label{tbl:t5}
	\begin{tabular}{l|l||p{1cm}|p{1cm}|p{1cm}|p{1cm} }							
		\hline	
		Colour & Metrics &TPR &SPC&PPV& NPV\\
		\hline	
		\multirow{2}{3.5em}{Image a}&pre-active&0.48&0.68&0.58&0.59\\
		&active&0.63&0.74&0.69&0.68\\
		\hline
		\multirow{2}{3.5em}{Image b}&pre-active& 0.69&0.75&0.72&0.72\\
		&active&0.76&0.80&0.78&0.78\\
		\hline
		\multirow{2}{3.5em}{Image c}&pre-active& 0.58&0.64&0.60&0.62\\
		&active&0.95&1.00&1.00&	0.95\\
		\hline
	\end{tabular}
	\end{table}

\subsubsection{Other experiments}

\begin{figure}
	\centering
	\includegraphics[width=0.34\linewidth]{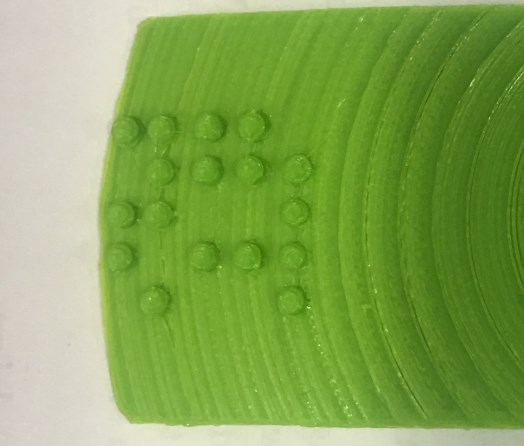} \hspace{1cm}
	\includegraphics[width=0.33\linewidth]{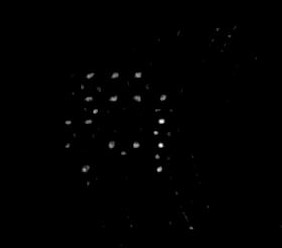} \vskip 0.5cm
	\includegraphics[width=0.34\linewidth]{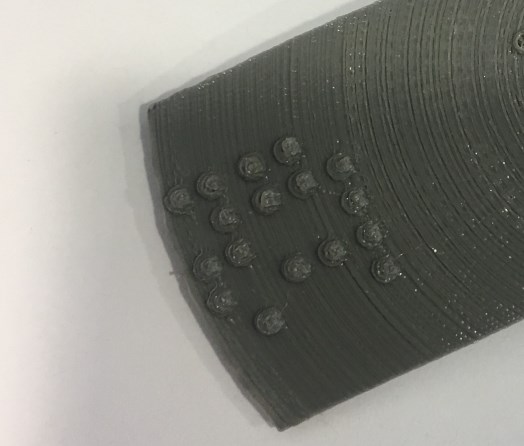} \hspace{1cm}
	\includegraphics[width=0.33\linewidth]{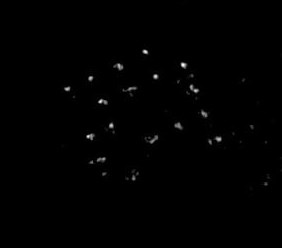}
	\caption{Confidence maps from non-planar watermarked surfaces.}
	\label{fig:non-planar}
\end{figure}

The current retrieval algorithm works on planar only surface watermarks. An extension 
of the approach to other watermark carrier surfaces seems to be a problem akin 
to surface reconstruction, and rather orthogonal to the machine learning part of 
the algorithm on which this paper concentrates. To test this claim we printed 
similar watermarks on spherical surfaces and processed their images with CNN-3DW. 
In the examples shown in Figure~\ref{fig:non-planar} we notice that good quality 
confidence maps are indeed constructed. 

In various watermarking applications the requirements for high retrieval accuracy are satisfied 
via the use of error correcting codes. In such cases, for obtaining optimal results, 
the choice of error correcting scheme could be informed by some knowledge of the 
distribution of the error. In an application scenario like ours, a natural question to ask 
is whether the position of a bit inside the $20\times 20$ information matrix, in 
particular its distance from the center of the matrix, correlates with the probability 
of correct retrieval. In a final test, on each of the 400 bit positions, we 
computed the accuracy rate over all 240 images of Experiment 1. We found a weak but not negligible 
correlation between accuracy rate and distance from the center of the information matrix with a 
Pearson coefficient of 0.46. That is, the further away from the center of the information 
matrix a bit is, the higher the probability to be correctly retrieved. While we do not 
have an intuitive explanation for this result, we notice that the algorithm performs better 
over the less congested parts of the watermark, that is, near the boundary and near the corners. 

	
	\section{Conclusions and Future Work} 
\label{sec:conclusions} 

We presented a method for retrieving digital information embedded as 
planar surface texture on a 3D printed object. The proposed method 
was tested extensively and it was shown to be a technically viable 
alternative to QR codes. Moreover, by integrating the information embedding 
process into the manufacturing process, we address several security 
vulnerabilities of the current widespread practice of displaying QR codes 
printed as separate items. 

\subsection{Limitations and Future Work} In tailoring our algorithm we 
imposed various limitations to the solution, most of them aiming at isolating 
for study the machine learning part of the watermark extraction pipeline. 
Thus, the registration of the confidence map relies on the use of a second 
printing material of distinct colour to mark the corners of a square region 
of interest. Moreover, as a convenient robust solution to the watermark orientation 
problem, we use four distinct corner symbols consisting of one, two, three and 
four small dots, respectively. In another limitation, we restricted ourselves to 
planar surfaces. While CNN-3DW generates good quality confidence maps 
of non-planar watermarked surfaces, the extraction of the watermark from the 
confidence maps is essentially a surface reconstruction problem, quite different 
in nature from the bump detection part. 

In the future, we plan to extend the current work into watermarks embedded 
on non-planar surfaces. As we discuss in Section~\ref{sec:experiments}, that 
extension shouldn't require any significant modifications of the CNN-3DW 
presented in this paper, but rather the development of new methods for 
processing the confidence maps. In another future research direction, 
we plan to develop higher capacity watermarks making use of more complex 
surface textures as carriers. In this case, we expect that the accurate 
retrieval of the watermark will require the development of more complex 
machine vision tools.  
\bibliographystyle{splncs}
\bibliography{reference}

\begin{thebibliography}{10}

\bibitem{thompson2016}
Thompson, M.K., Moroni, G., Vaneker, T., Fadel, G., Campbell, R.I., Gibson, I.,
  Bernard, A., Schulz, J., Graf, P., Ahuja, B.,  et~al.:
\newblock Design for additive manufacturing: Trends, opportunities,
  considerations, and constraints.
\newblock CIRP annals \textbf{65} (2016)  737--760

\bibitem{wong2012review}
Wong, K.V., Hernandez, A.:
\newblock A review of additive manufacturing.
\newblock ISRN Mechanical Engineering \textbf{2012} (2012)

\bibitem{terry2012additive}
Terry, W.:
\newblock Additive manufacturing and 3d printing state of the industry.
\newblock Annual Worldwide Progress Report, Wohlers Associations (2012)

\bibitem{ohbuchi02}
Ohbuchi, R., Mukaiyama, A., Takahashi, S.:
\newblock A frequency-domain approach to watermarking 3{D} shapes.
\newblock Computer Graphics Forum \textbf{21} (2002)  373--382

\bibitem{bors2006}
Bors, A.G.:
\newblock Watermarking mesh-based representations of 3-d objects using local
  moments.
\newblock IEEE Transactions on Image processing \textbf{15} (2006)  687--701

\bibitem{luo11}
Luo, M., Bors, A.G.:
\newblock Surface-preserving robust watermarking of 3{D} shapes.
\newblock IEEE Trans. on Image Processing \textbf{20} (2011)  2813--2826

\bibitem{yang17}
Yang, Y., Pintus, R., Rushmeier, H., Ivrissimtzis, I.:
\newblock A 3d steganalytic algorithm and steganalysis-resistant watermarking.
\newblock IEEE transactions on visualization and computer graphics \textbf{23}
  (2017)  1002--1013

\bibitem{macq15}
Macq, B., Alface, P.R., Montanola, M.:
\newblock Applicability of watermarking for intellectual property rights
  protection in a 3{D} printing scenario.
\newblock In: {Proc. of the International Conference on 3D Web Technology}, ACM
  (2015)  89--95

\bibitem{hou15}
Hou, J.U., Kim, D.G., Choi, S., Lee, H.K.:
\newblock {3D Print-Scan Resilient Watermarking Using a Histogram-Based
  Circular Shift Coding Structure}.
\newblock In: Proc. of the ACM Workshop on Information Hiding and Multimedia
  Security. (2015)  115--121

\bibitem{lecun2015deep}
LeCun, Y., Bengio, Y., Hinton, G.:
\newblock Deep learning.
\newblock nature \textbf{521} (2015)  436

\bibitem{lecun2010convolutional}
LeCun, Y., Kavukcuoglu, K., Farabet, C.,  et~al.:
\newblock Convolutional networks and applications in vision.
\newblock In: ISCAS. Volume 2010. (2010)  253--256

\bibitem{robert2014machine}
Robert, C.:
\newblock Machine learning, a probabilistic perspective (2014)

\bibitem{krizhevsky2012imagenet}
Krizhevsky, A., Sutskever, I., Hinton, G.E.:
\newblock Imagenet classification with deep convolutional neural networks.
\newblock In: Advances in neural information processing systems. (2012)
  1097--1105

\bibitem{simonyan2014very}
Simonyan, K., Zisserman, A.:
\newblock Very deep convolutional networks for large-scale image recognition.
\newblock arXiv:1409.1556 (2014)

\bibitem{szegedy2015going}
{Szegedy, C. et al.}:
\newblock Going deeper with convolutions.
\newblock In: IEEE CVPR. (2015)

\bibitem{zhang2015cross}
Zhang, C., Li, H., Wang, X., Yang, X.:
\newblock Cross-scene crowd counting via deep convolutional neural networks.
\newblock In: Proceedings of the IEEE Conference on Computer Vision and Pattern
  Recognition. (2015)  833--841

\bibitem{zhang2016single}
Zhang, Y., Zhou, D., Chen, S., Gao, S., Ma, Y.:
\newblock Single-image crowd counting via multi-column convolutional neural
  network.
\newblock In: IEEE CVPR. (2016)  589--597

\bibitem{mundhenk2016large}
Mundhenk, T.N., Konjevod, G., Sakla, W.A., Boakye, K.:
\newblock A large contextual dataset for classification, detection and counting
  of cars with deep learning.
\newblock In: European Conference on Computer Vision, Springer (2016)  785--800

\bibitem{ojala2000gray}
Ojala, T., Pietik{\"a}inen, M., M{\"a}enp{\"a}{\"a}, T.:
\newblock Gray scale and rotation invariant texture classification with local
  binary patterns.
\newblock In: ECCV, Springer (2000)  404--420

\bibitem{zeiler2013rectified}
{Zeiler M. et al.}:
\newblock On rectified linear units for speech processing.
\newblock In: Acoustics, Speech and Signal Processing (ICASSP), IEEE (2013)
  3517--3521

\bibitem{chollet2015keras}
Chollet, F.,  et~al.:
\newblock Keras.
\newblock \url{https://github.com/fchollet/keras} (2015)

\bibitem{kingma2014adam}
Kingma, D.P., Ba, J.:
\newblock Adam: A method for stochastic optimization.
\newblock arXiv preprint arXiv:1412.6980 (2014)

\bibitem{brown1992survey}
Brown, L.G.:
\newblock A survey of image registration techniques.
\newblock ACM computing surveys (CSUR) \textbf{24} (1992)  325--376

\bibitem{otsu1979threshold}
Otsu, N.:
\newblock A threshold selection method from gray-level histograms.
\newblock IEEE transactions on systems, man, and cybernetics \textbf{9} (1979)
  62--66

\end{thebibliography}
	
\end{document}